\documentclass{article}
\usepackage{spconf,amsmath,amssymb,amsthm,graphicx,xurl,hyperref}
\usepackage{algorithm,algorithmic}
\usepackage[table]{xcolor}
\usepackage{booktabs}
\definecolor{tabgreen}{RGB}{58,181,74}
\usepackage{bm}
\usepackage{float}   

\DeclareMathOperator{\tr}{tr}
\newtheorem{proposition}{Proposition}
\newtheorem*{propositionrestatement}{Proposition}
\newcommand{\knn}{k\text{NN}}
\newcommand{\supp}{the supplementary material}

\title{ELEMENT: Episodic and Lifelong Exploration via Maximum Entropy}
\name{Hongming Li$^{1,5,\dagger}$, Zhao Yang$^{2,\dagger}$, Xiaoxuan Liang$^3$, Shujian Yu$^{2,\star}$, Jose C. Principe$^{4,5}$
\thanks{$^{\dagger}$Equal contribution. $^{\star}$Corresponding author.}}
\address{$^{1}$E-surfing Digital Life Technology Co., Ltd., China Telecom, Shanghai, China; \\
$^{2}$Vrije Universiteit Amsterdam, The Netherlands; 
$^{3}$University of Massachusetts Amherst, USA;\\
$^{4}$INESC TEC, University of Porto, Portugal; 
$^{5}$University of Florida, USA}
\begin{document}
\ninept
\maketitle
\begin{abstract}
Reinforcement learning agents depend on reward signals whose density is rarely under the designer's control, and when such signals are absent, an agent must generate its own drive to explore. State entropy maximization offers a principled objective for this, but existing methods break down at scale in two ways: the intrinsic reward vanishes once a state has been visited, discouraging revisits to the very gateways that lead onward, and estimating entropy over millions of accumulated observations becomes computationally prohibitive. We address both with Episodic and Lifelong Exploration via Maximum Entropy (ELEMENT), a multiscale intrinsically motivated framework for reward-free exploration that transfers to downstream tasks. ELEMENT couples lifelong entropy maximization with a complementary episodic term acting on a faster timescale. For the episodic term, we derive average episodic state entropy, an intrinsic reward that is the exact minimizer of a tractable upper bound on the reward-decomposition objective; for the lifelong term, we propose a $k$NN graph-based estimator that keeps entropy tractable without forgetting. ELEMENT consistently outperforms state-of-the-art intrinsic reward baselines on state coverage and unsupervised pre-training. Videos, code, and supplementary material: \url{https://sites.google.com/view/element-rl}.
\end{abstract}
\begin{keywords}
Reinforcement learning, exploration, state entropy maximization, intrinsic reward, $k$NN graph
\end{keywords}
\section{Introduction}
\label{sec:intro}
Reinforcement learning (RL) has achieved remarkable success in games and simulated robotics, yet its real-world use remains limited because the density of reward signals is highly variable and beyond the user's control. When extrinsic rewards are sparse or missing, intrinsically motivated exploration supplies a substitute: an internally generated reward that encourages the agent to visit novel and diverse states.

State entropy $H(\mathbf{s})$ has emerged as a well-defined measure of the diversity of visited states. It serves as a popular intrinsic reward~\cite{liu2021behavior,mutti2021task,seo2021state,yuan2022renyi} and as the basis of sampling methods~\cite{hazan2019provably,zhang2021exploration,yarats2021reinforcement,tiapkin2023fast}. In non-tabular settings, APT~\cite{liu2021behavior} estimates it from $k$-nearest-neighbor ($k$NN) distances, i.e., the Kozachenko-Leonenko estimator~\cite{singh2003nearest}. Such distance-based methods~\cite{liu2021behavior,seo2021state,yuan2022renyi} share a weakness: their lifelong reward vanishes. Once a state has been visited, its novelty decays sharply, and the agent is discouraged from returning even when that state is a gateway to unexplored regions~\cite{bellemare2016unifying,ecoffet2019go,badia2020never}. Parametric curiosity methods~\cite{pathak2017curiosity,burda2018exploration} measure novelty differently, through the prediction error of an auxiliary model, and P2E~\cite{sekar2020planning} through the disagreement of a model ensemble, but they pay for it with extra models and hyper-parameters; NGU~\cite{badia2020never} and Go-Explore~\cite{ecoffet2019go} add episodic mechanisms on top of such models (see \supp).

The episodic idea does not require a parametric model. Episodic novelty encourages an agent to revisit states that are familiar but not yet fully explored~\cite{badia2020never,ecoffet2019go}, exactly what a vanishing lifelong reward forbids. Our approach, ELEMENT, therefore fosters Episodic and Lifelong Exploration jointly under the Maximum Entropy principle: the lifelong component slowly penalizes revisiting states that are already well represented in the global visitation history, while the episodic component rapidly penalizes repetition within an episode. The two operate on complementary timescales, and both are computed non-parametrically from state entropy.

Turning this idea into an algorithm raises two difficulties. First, the entropy of an episode can only be computed at its conclusion, which yields a sparse trajectory-level reward that harms sample efficiency and learning stability~\cite{rengarajan2022reinforcement}. We resolve this by deriving average episodic state entropy, a proxy reward that decomposes the trajectory-level bonus into dense per-step rewards and provably minimizes a tractable upper bound of the reward-decomposition objective. Second, in the lifelong setting the set of visited states grows without bound. We show that the $k$NN and kernel density estimators admit exact or $\varepsilon$-accurate sum decompositions over $k$NN quantities, while the matrix-based R\'enyi functional~\cite{giraldo2014measures} is a strictly monotone function of such a sum. This shared structure lets us organize the memory in a fast $k$NN graph, combining greedy graph search~\cite{hajebi2011fast} with online updating~\cite{debatty2016fast}, for scalable lifelong reward estimation.

In summary, our contributions are: (1)~ELEMENT, a non-parametric exploration method that jointly maximizes episodic and lifelong state entropy, yielding multiscale intrinsic rewards without any extrinsic signal; (2)~average episodic state entropy, a Markovian proxy reward with an optimality guarantee with respect to a tractable upper bound; (3)~a theoretical characterization relating three popular entropy estimators to $k$NN quantities, which enables scalable $k$NN-graph-based lifelong rewards; and (4)~experiments on five environments showing that ELEMENT outperforms four state-of-the-art baselines in episodic entropy maximization, state coverage, and unsupervised pre-training for downstream tasks.

\section{Background}
\label{sec:background}
\subsection{Entropy estimators}
For $x$ with density $p(x)$, R\'enyi's $\alpha$-entropy~\cite{renyi1961measures} is $H^{\alpha}(x)=\frac{1}{1-\alpha}\log_2\!\int p^{\alpha}(x)dx$, which recovers Shannon entropy as $\alpha\to1$. Given $N$ samples $\{\mathbf{s}_i\}_{i=1}^{N}$ and a Gram matrix $K(i,j)=\kappa(\mathbf{s}_i,\mathbf{s}_j)$ with a Gaussian kernel $\kappa_\sigma(\mathbf{s}_i,\mathbf{s}_j)=\exp(-\|\mathbf{s}_i-\mathbf{s}_j\|_2^2/2\sigma)$, the matrix-based functional~\cite{giraldo2014measures} estimates it as
\begin{equation}\label{eq:renyi}
H_{\alpha}(\mathbf{s})=\tfrac{1}{1-\alpha}\log_2\tr(A^{\alpha})=\tfrac{1}{1-\alpha}\log_2\textstyle\sum_{i=1}^{N}\lambda_i(A)^{\alpha},
\end{equation}
where $A=K/\tr(K)$ and $\lambda_i(A)$ are its eigenvalues; as $\alpha\to1$ this gives $H_1(\mathbf{s})=-\sum_i\lambda_i(A)\log_2\lambda_i(A)$. We also use the kernel density estimator (KDE),
\begin{equation}\label{eq:kde}
H_{\text{KDE}}(\mathbf{s})=-\tfrac{1}{N}\textstyle\sum_{i=1}^{N}\log\!\big[\tfrac{1}{N}\sum_{j=1}^{N}\kappa(\mathbf{s}_i,\mathbf{s}_j)\big],
\end{equation}
and the $k$NN estimator~\cite{singh2003nearest} is defined through neighbor distances. Let $\mathbf{s}_i^{(k)}$ be the $k$-th nearest neighbor of $\mathbf{s}_i$ and $\rho_i:=\|\mathbf{s}_i-\mathbf{s}_i^{(k)}\|_2$; then
\begin{equation}\label{eq:knn}
H_{\knn}(\mathbf{s})=\tfrac{1}{N}\textstyle\sum_{i=1}^{N}\log\frac{N\rho_i^{d}\pi^{d/2}}{k\,\Gamma(d/2+1)}+C_k,
\end{equation}
where $C_k=\log k-\Psi(k)$ is a bias correction, $\Psi$ is the digamma function, and $d$ is the state dimensionality.

\subsection{Exploration with maximum state entropy}
An MDP~\cite{sutton1998reinforcement} is a tuple $(\mathcal{S},\mathcal{A},p,r,\gamma)$; an episode $\tau=\{\mathbf{s}_1,\dots,\mathbf{s}_T\}$ is induced by a policy $\pi$, which maximizes the discounted return $J(\pi)=\mathbb{E}_{\tau\sim\pi}\big[\sum_{t}\gamma^{t} r(\mathbf{s}_t,\mathbf{a}_t)\big]$. Without extrinsic rewards, a natural objective is $\arg\max_\pi H(p_\pi(\mathbf{s}))$, where $p_\pi$ is the state distribution induced by $\pi$~\cite{hazan2019provably}; follow-ups include a R\'enyi variant~\cite{zhang2021exploration}, reduced sample complexity~\cite{tiapkin2023fast}, and finite-horizon settings~\cite{mutti2021task}. A simpler and widely adopted approach is APT~\cite{liu2021behavior}, which uses $k$NN distances directly as intrinsic rewards,
\begin{equation}\label{eq:apt}
r^i(\mathbf{s}):=\log(\|\mathbf{s}-\mathbf{s}^{(k)}\|_2+1),
\end{equation}
where $\mathbf{s}^{(k)}$ is the $k$-th nearest neighbor of $\mathbf{s}$ among previously visited states. Variants differ in representation learning: contrastive features~\cite{liu2021behavior}, prototypes~\cite{yarats2021reinforcement}, random encoders (RE3)~\cite{seo2021state}, and R\'enyi entropy (RISE)~\cite{yuan2022renyi}. All operate purely in a lifelong setting and are thus exposed to the vanishing reward problem.

\section{Intrinsic Reward with ELEMENT}
\label{sec:method}
We consider a fully unsupervised setting without extrinsic rewards. ELEMENT combines two rewards that act on different timescales: an episodic reward $r_{ep}$, which rapidly encourages diverse states within an episode by maximizing $H_{\mathbf{s}\in\tau}(\mathbf{s})$ over $\tau=\{\mathbf{s}_t\}_{t=1}^{T_\tau}$, and a lifelong reward $r_l$, which slowly discourages revisits by maximizing $H_{\mathbf{s}\in D}(\mathbf{s})$ over a global memory $D=\{\mathbf{s}_i\}_{i=1}^{N}$. The two are combined as
\begin{equation}\label{eq:reward}
r^i(\mathbf{s})=r_{ep}(\mathbf{s})+\beta\, r_l(\mathbf{s}),
\end{equation}
with $\beta>0$ weighting the lifelong term. Both modules share a fixed encoder $f:\mathcal{O}\to\mathcal{S}$ that maps observations to $d$-dimensional states. The encoder must be fixed: a drifting one would let the agent visit spurious ``states'' without meaningful actions, corrupting the entropy signal. We use a fixed random encoder for MuJoCo, which is sufficient for state entropy estimation~\cite{seo2021state}, and a Spatiotemporal Deep Infomax encoder~\cite{anand2019unsupervised} for Mario.

\begin{figure}[t]
\centering
\includegraphics[width=\linewidth]{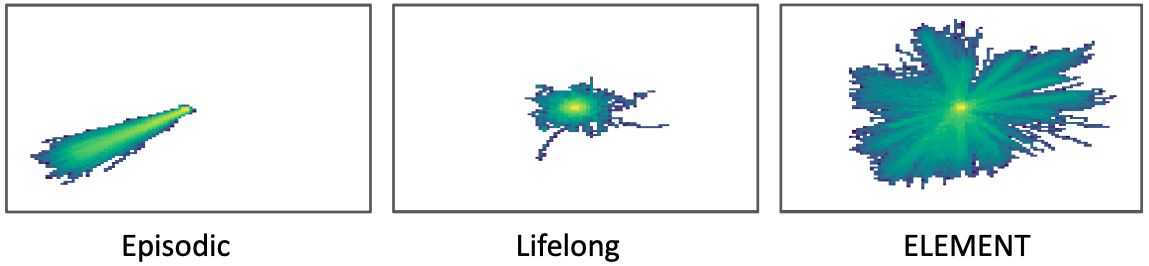}
\caption{State visitation heatmaps on Ant under different rewards (lighter = more visits). ELEMENT combines the merits of episodic and lifelong rewards, exploring radially in all directions.}
\label{fig:ant}
\end{figure}

Fig.~\ref{fig:ant} shows why the two rewards are complementary, using the MuJoCo Ant, which starts at the center of a 2D plane. With $r_{ep}$ alone, the agent commits to a single direction, since that maximizes diversity within an episode; with $r_l$ alone, it avoids past directions but never explores any of them in depth. ELEMENT integrates both and explores radially outward like ``fireworks''.

\subsection{Episodic intrinsic reward}
\label{sec:episodic}
The difficulty with the episodic term is that its entropy is defined over whole episodes, so the agent receives a single sparse feedback $R_{ep}(\tau)=H_{\mathbf{s}\in\tau}(\mathbf{s})$ only when an episode ends. Following trajectory-wise reward redistribution~\cite{gangwani2020learning,ren2021learning}, we assume a dense reward $r_{ep}(\mathbf{s})$ with $R_{ep}(\tau)\approx\sum_{t=0}^{T_\tau-1}r_{ep}(\mathbf{s}_{t+1})$, so that
\begin{equation}\label{eq:traj}
J_{ep}(\pi)=\mathbb{E}_{\tau\sim\pi}\big[H_{\mathbf{s}\in\tau}(\mathbf{s})\big]\approx\mathbb{E}_{\tau\sim\pi}\Big[\textstyle\sum_{t=0}^{T_\tau-1}r_{ep}(\mathbf{s}_{t+1})\Big].
\end{equation}
A common route is to parameterize $r_{ep}(\cdot;\theta)$ and minimize~\cite{ren2021learning}
\begin{equation}\label{eq:loss}
\mathcal{L}(r_{ep})=\mathbb{E}_{\tau\in D}\Big[\big(H_{\mathbf{s}\in\tau}(\mathbf{s})-\textstyle\sum_{t=0}^{T_\tau-1}r_{ep}(\mathbf{s}_{t+1};\theta)\big)^2\Big],
\end{equation}
but this adds hyper-parameters and tuning. Inspired by episode-space smoothing~\cite{gangwani2020learning}, we avoid the parametric model altogether and define the non-parametric average episodic state entropy
\begin{equation}\label{eq:practical}
\begin{aligned}
r_{ep}(\mathbf{s})&:=\mathbb{E}_{\tau\in D}\big[H_{\mathbf{s}_i\in\tau}(\mathbf{s}_i)\,\big|\,\mathbf{s}\in\tau\big]\\
&=\frac{\mathbb{E}_{\tau\in D}[H_{\mathbf{s}_i\in\tau}(\mathbf{s}_i)\,\mathbb{I}(\mathbf{s}\in\tau)]}{\mathbb{E}_{\tau\in D}[\mathbb{I}(\mathbf{s}\in\tau)]},
\end{aligned}
\end{equation}
where $\mathbb{I}(\mathbf{s}\in\tau)$ indicates whether $\tau$ contains $\mathbf{s}$. States that consistently appear in high-entropy episodes thus receive larger rewards. In continuous spaces each exact state typically appears in a single episode, so Eq.~(\ref{eq:practical}) assigns to every state the entropy of its own episode (Algorithm~\ref{alg:element}); the conditional form matters when states recur, e.g., in tabular settings. Any estimator of Sec.~\ref{sec:background} can be plugged in. Moreover, Eq.~(\ref{eq:practical}) is optimal for an upper bound of Eq.~(\ref{eq:loss}).

\begin{proposition}\label{pro:optimal}
Assume all episodes in $D$ have length $T$ and each state occurs at most once per episode. Define the Monte-Carlo upper bound~\cite{ren2021learning}
$\tilde{\mathcal{L}}(r_{ep})=\mathcal{L}(r_{ep})+\mathbb{E}_{\tau\in D}[T^2\,V_{\mathbf{s}\in\tau}(r_{ep}(\mathbf{s}))]\geq\mathcal{L}(r_{ep})$,
where $V_{\mathbf{s}\in\tau}(\cdot)\geq0$ is the within-episode reward variance. The unique minimizer of $\tilde{\mathcal{L}}$ is
\begin{equation}\label{eq:optimal}
r^{*}_{ep}(\mathbf{s})=\tfrac{1}{T}\,\mathbb{E}_{\tau\in D}\big[H_{\mathbf{s}_i\in\tau}(\mathbf{s}_i)\,\big|\,\mathbf{s}\in\tau\big].
\end{equation}
\end{proposition}
\noindent Eqs.~(\ref{eq:practical}) and (\ref{eq:optimal}) differ only by the constant $T$, which is absorbed by reward normalization. Optimality is with respect to $\tilde{\mathcal{L}}$ rather than $\mathcal{L}$: the gap penalizes within-episode reward variation, so the proxy deliberately assigns credit across rather than within episodes. Proposition~\ref{pro:optimal} concerns the undiscounted sum in Eq.~(\ref{eq:traj}); with $\gamma$ close to 1, as in our experiments, the mismatch with the discounted return is small. The proof is given in \supp.
When episode lengths $T_\tau$ differ, defining $\tilde{\mathcal{L}}$ over transitions sampled uniformly from the buffer (line~14 of Algorithm~\ref{alg:element}) yields a $T_\tau$-weighted version of Eq.~(\ref{eq:practical}); see \supp.

\subsection{Lifelong intrinsic reward}
The sum decomposition of Eq.~(\ref{eq:traj}) does not carry over to the lifelong setting, where no estimator can handle tens of millions of visited states at once. We instead seek per-state rewards $r_l(\mathbf{s}_i)$ whose sum, or a strictly monotone function thereof, tracks $H_{\mathbf{s}\in D}(\mathbf{s})$ up to constants. Let $\knn(i)$ be the index set of the $k$ nearest neighbors of $\mathbf{s}_i$ (excluding $i$), $S_i:=\sum_{j\in\knn(i)}\kappa(\mathbf{s}_i,\mathbf{s}_j)$ and $S'_i:=\sum_{j\in\knn(i)}\kappa^2(\mathbf{s}_i,\mathbf{s}_j)$.

\begin{proposition}\label{pro:intrinsic}
Given $N$ states, a Gaussian kernel $\kappa_\sigma$, $\rho_i$ as in Eq.~(\ref{eq:knn}), and a tolerance $\varepsilon\in(0,1)$:
\begin{align}
\text{(i)}\;\; & H_{\knn}(\mathbf{s})=\tfrac{d}{N}\textstyle\sum_{i=1}^{N}\log\rho_i+c_{N,k,d},\label{eq:rel-i}\\
\text{(ii)}\;\; & \Big|H_{\text{KDE}}(\mathbf{s})-\big(\log N-\tfrac{1}{N}\textstyle\sum_{i=1}^{N}\log(1+S_i)\big)\Big|\leq\varepsilon\notag\\
& \quad\text{if }\rho_i\geq\big[2\sigma\log\tfrac{N-k-1}{\varepsilon}\big]^{1/2}\;\forall i,\label{eq:rel-ii}\\
\text{(iii)}\;\; & H_2(\mathbf{s})=2\log_2N-\log_2\big(N+\textstyle\sum_{i=1}^{N}S'_i\big)+\delta,\notag\\
& \quad|\delta|\leq\tfrac{\varepsilon}{\ln2}\;\text{ if }\rho_i\geq\big[\sigma\log\tfrac{N-k-1}{\varepsilon}\big]^{1/2}\;\forall i,\label{eq:rel-iii}
\end{align}
where $c_{N,k,d}=\log N+\log\frac{\pi^{d/2}}{k\Gamma(d/2+1)}+C_k$. Relation (i) is exact; (ii) and (iii) exploit the exponential decay of $\kappa_\sigma$, and since $\kappa^2$ has bandwidth $\sigma/2$, the condition in (iii) is milder than in (ii) by a factor $\sqrt2$.
\end{proposition}
\noindent Relations (i) and (ii) are additive: the summed return of a per-state reward is affine in the entropy estimate. Relation (iii) is not: $H_2$ is a strictly decreasing function of $\sum_iS'_i$, so per-state rewards $-S'_i$ increase $H_2$ monotonically, although the return is not proportional to it. The conditions in (ii) and (iii) also require the $k$NN radius to exceed a threshold growing as $\sqrt{\sigma\log N}$, whereas empirical $k$NN distances shrink as states accumulate. The exact relation (i) is immune to this, which is why we build the lifelong module on the $k$NN estimator. The proof is in \supp.

Previous $k$NN-based methods~\cite{liu2021behavior,seo2021state,yuan2022renyi} implement relation (i) but cope with the growing memory via first-in-first-out queues or subsampling, which defeats the lifelong goal through rapid ``forgetting''. We instead organize visited states in a $k$NN graph $\mathcal{G}=(D_{\mathcal{G}},\mathcal{E})$, $D_{\mathcal{G}}\subseteq D$, where $\mathbf{s}_i\to\mathbf{s}_j$ if $\mathbf{s}_j$ is among the $k$ nearest neighbors of $\mathbf{s}_i$ in $D_{\mathcal{G}}$. Unlike a queue, $D_{\mathcal{G}}$ accumulates states over the entire training history. Let $\text{GNNS}(\mathcal{G},\mathbf{s})$ denote graph nearest neighbor search~\cite{prokhorenkova2020graph} returning the $k$ nearest nodes of $\mathbf{s}$, and $\mathbf{s}^{(k)}_{\mathcal{G}}$ the $k$-th of these. The lifelong reward is then
\begin{equation}\label{eq:gnn-reward}
r_l(\mathbf{s}):=\log(\|\mathbf{s}-\mathbf{s}^{(k)}_{\mathcal{G}}\|_2+1),
\end{equation}
a monotone surrogate of the per-state term $d\log\rho$ in relation (i), where the $+1$ stabilizes the reward near zero distance as in APT~\cite{liu2021behavior}.

In practice, Fast Approximate GNN Search~\cite{hajebi2011fast} and Fast Online $k$NN Graph Building~\cite{debatty2016fast} work well: search starts from $R_2$ random nodes and, for at most $R_1$ steps, greedily moves to the neighbor closest to the query; insertion uses the neighbors of a new point as starting points to find, up to depth $h$, existing nodes for which the new point is now a nearest neighbor (pseudo-code in \supp). A reward query thus costs $\mathcal{O}(R_1R_2kd)$ distance evaluations and an insertion $\mathcal{O}(k^{h+1}d)$, so a run with $N$ queries and $N_{\mathcal{G}}$ insertions costs $\mathcal{O}(NR_1R_2kd+N_{\mathcal{G}}k^{h+1}d)$, versus $\mathcal{O}(N^2d)$ for brute-force search over the full history.

\begin{algorithm}[t]
\caption{ELEMENT for off-policy RL}
\label{alg:element}
\begin{algorithmic}[1]
\STATE \textbf{Init:} policy $\pi$, $k$NN graph $\mathcal{G}$, buffer $D\leftarrow\emptyset$, $t_{\rm env}\leftarrow0$, batch size $B$, graph update interval $U$, duration $T_u$, weight $\beta$
\FOR{each episode $\tau$}
  \STATE $D_\tau\leftarrow\emptyset$
  \FOR{$t=1,\dots,T_\tau$}
    \STATE $\mathbf{a}_t\sim\pi(\cdot|\mathbf{s}_t)$, $\mathbf{s}_{t+1}\sim P(\cdot|\mathbf{s}_t,\mathbf{a}_t)$, $t_{\rm env}\leftarrow t_{\rm env}+1$
    \STATE $D_\tau\leftarrow D_\tau\cup\{(\mathbf{s}_t,\mathbf{a}_t,\mathbf{s}_{t+1})\}$
    \IF{$0<(t_{\rm env}\bmod U)<T_u$}
      \STATE $\mathcal{G}\leftarrow\mathcal{G}\cup\{\mathbf{s}_{t+1}\}$ \COMMENT{online graph update}
    \ENDIF
  \ENDFOR
  \STATE $R_{ep}(\tau)\leftarrow H(\mathbf{s}\in D_\tau)$ \COMMENT{episodic entropy}
  \STATE $D\leftarrow D\cup\{(\mathbf{s}_t,\mathbf{a}_t,\mathbf{s}_{t+1},R_{ep}(\tau))\}_{t=1}^{T_\tau}$
  \FOR{each policy update step}
    \STATE Sample $\{\mathbf{s}^{(n)},\mathbf{a}^{(n)},\mathbf{s}'^{(n)},r^{(n)}_{ep}\}_{n=1}^{B}\sim D$
    \STATE Compute $r_l$ via Eq.~(\ref{eq:gnn-reward}); normalize $r_{ep}$, $r_l$
    \STATE $r^i\leftarrow\hat r_{ep}+\beta\hat r_l$; update $\pi$ with $r^i$
  \ENDFOR
\ENDFOR
\end{algorithmic}
\end{algorithm}

Algorithm~\ref{alg:element} summarizes the resulting off-policy pipeline. On-policy algorithms are accommodated by recording the episodic entropy of each state and assigning to $r_{ep}(\mathbf{s})$ the average recorded entropy of the $k$ nearest neighbors of $\mathbf{s}$. Since lifelong exploration should only slowly discourage revisits, the graph is updated every $U$ environment steps for $T_u$ steps; between updates $r_l$ is stationary.

\section{Experiments}
\label{sec:exp}
\begin{figure}[H]
\centering
\includegraphics[width=\linewidth]{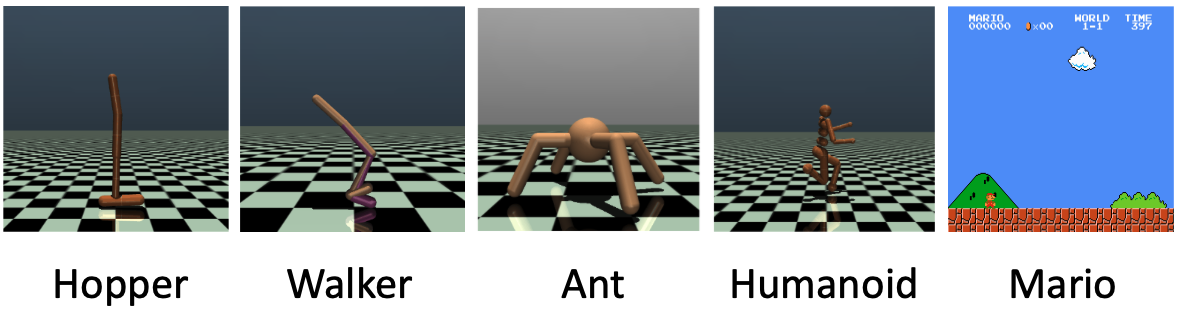}
\caption{The five evaluation tasks: four MuJoCo continuous-control environments and Mario with a discrete action space.}
\label{fig:envs}
\end{figure}

We evaluate on four MuJoCo continuous-control tasks (Fig.~\ref{fig:envs}), Hopper (11-D observations), Walker (17-D), Ant (27-D), and Humanoid (376-D), and on Mario, a 2D platformer with 16 discrete actions in which the agent starts at the leftmost point of Stage 1-1. We compare against four state-of-the-art baselines: RND~\cite{burda2018exploration}, NGU~\cite{badia2020never}, RE3~\cite{seo2021state}, and RISE~\cite{yuan2022renyi} (intrinsic reward only). All methods use SAC~\cite{haarnoja2018soft2} on MuJoCo and A3C~\cite{mnih2016asynchronous} on Mario, with $\beta=0.5$ after normalizing $r_{ep}$ and $r_l$ to $[0,1]$. Encoders, hyper-parameters, trajectory and reward-distribution visualizations, and experiments augmenting extrinsic rewards are provided in \supp.

\begin{table}[!tb]
\centering\ninept
\setlength{\tabcolsep}{2.1pt}
\caption{Final episodic state entropy (mean\,$\pm$\,std over runs) at the last checkpoint of each environment. Darker green indicates higher entropy; column-wise best in bold.}
\label{tab:final_entropy}
\begin{tabular}{lccccc}
\toprule
Method & Hopper & Walker2D & Ant & Humanoid & Mario \\
\midrule
\textbf{ELEMENT} & \cellcolor{tabgreen!52}\textbf{6.67}\,\tiny{$\pm$0.30} & \cellcolor{tabgreen!52}\textbf{6.25}\,\tiny{$\pm$0.17} & \cellcolor{tabgreen!52}\textbf{7.43}\,\tiny{$\pm$0.69} & \cellcolor{tabgreen!52}\textbf{5.33}\,\tiny{$\pm$0.37} & \cellcolor{tabgreen!52}\textbf{8.22}\,\tiny{$\pm$0.37} \\
RE3 & \cellcolor{tabgreen!33}4.60\,\tiny{$\pm$1.04} & \cellcolor{tabgreen!38}4.72\,\tiny{$\pm$0.16} & \cellcolor{tabgreen!27}4.38\,\tiny{$\pm$0.41} & \cellcolor{tabgreen!8}3.22\,\tiny{$\pm$0.52} & \cellcolor{tabgreen!35}7.24\,\tiny{$\pm$0.64} \\
RISE & \cellcolor{tabgreen!26}3.94\,\tiny{$\pm$0.63} & \cellcolor{tabgreen!39}4.85\,\tiny{$\pm$0.37} & \cellcolor{tabgreen!32}5.01\,\tiny{$\pm$0.20} & \cellcolor{tabgreen!39}4.69\,\tiny{$\pm$0.28} & \cellcolor{tabgreen!23}6.50\,\tiny{$\pm$0.33} \\
NGU & \cellcolor{tabgreen!17}2.97\,\tiny{$\pm$0.79} & \cellcolor{tabgreen!18}2.49\,\tiny{$\pm$0.60} & \cellcolor{tabgreen!19}3.42\,\tiny{$\pm$0.74} & \cellcolor{tabgreen!14}3.51\,\tiny{$\pm$0.13} & \cellcolor{tabgreen!8}5.64\,\tiny{$\pm$0.65} \\
RND & \cellcolor{tabgreen!8}1.97\,\tiny{$\pm$0.92} & \cellcolor{tabgreen!8}1.33\,\tiny{$\pm$0.31} & \cellcolor{tabgreen!8}2.03\,\tiny{$\pm$0.53} & \cellcolor{tabgreen!19}3.77\,\tiny{$\pm$0.45} & \cellcolor{tabgreen!16}6.12\,\tiny{$\pm$0.22} \\
\bottomrule
\end{tabular}
\end{table}
\begin{figure*}[!tb]
\centering
\includegraphics[width=\linewidth]{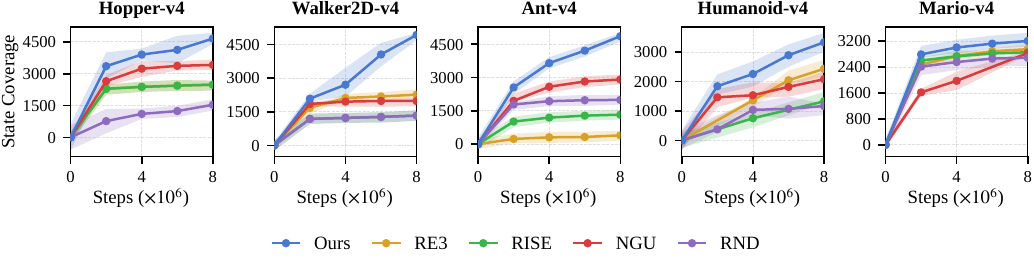}
\caption{State coverage: number of unique visited states (on a $100\times100$ grid of $x$-$y$ or $x$-$z$ coordinates) throughout training; mean over 5 runs.}
\label{fig:cov}
\end{figure*}
\begin{figure}[!tb]
\centering
\includegraphics[width=\linewidth]{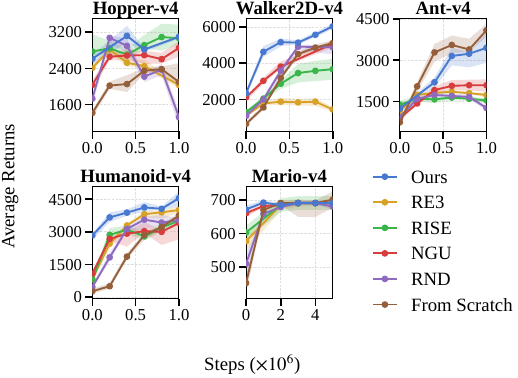}
\caption{Downstream task performance after unsupervised pre-training with ELEMENT and the baselines; mean over 5 runs.}
\label{fig:transfer}
\end{figure}

\textbf{Can ELEMENT maximize episodic state entropy?}
We set $\beta=0$ in Eq.~(\ref{eq:reward}) to isolate the episodic component and evaluate with Shannon entropy from Eq.~(\ref{eq:renyi}) ($\alpha=1.001$). Table~\ref{tab:final_entropy} reports the entropy at the last checkpoint: ELEMENT attains the highest value in every environment, by about one nat on Mario and more than two on Hopper and Ant; learning curves in \supp\ show it also reaches these levels earlier.

\textbf{Can ELEMENT explore new states efficiently?}
We measure coverage by discretizing the $x$-$y$ or $x$-$z$ coordinates into $100\times100$ histograms and counting unique visited cells during training. Using both rewards, ELEMENT achieves the highest coverage in every environment (Fig.~\ref{fig:cov}), and the gap widens with training: the baselines rely purely on lifelong novelty, so their growth slows as rewards vanish along visited routes, exactly the failure the episodic term counters.

\textbf{Does task-agnostic exploration help downstream tasks?}
After unsupervised pre-training we save 10 checkpoints per method, select the one with the highest task reward as initialization, and continue training with extrinsic rewards (Fig.~\ref{fig:transfer}). ELEMENT consistently improves downstream performance and accelerates learning relative to training from scratch. In Hopper, Walker, Humanoid, and Mario it enables zero-shot transfer, achieving strong task performance at step 0. Ant is the exception: all exploration agents settle into sub-optimal locomotion modes and are outperformed by SAC from scratch, though ELEMENT remains the best exploration baseline.

\begin{figure}[!tb]
\centering
\includegraphics[width=\linewidth]{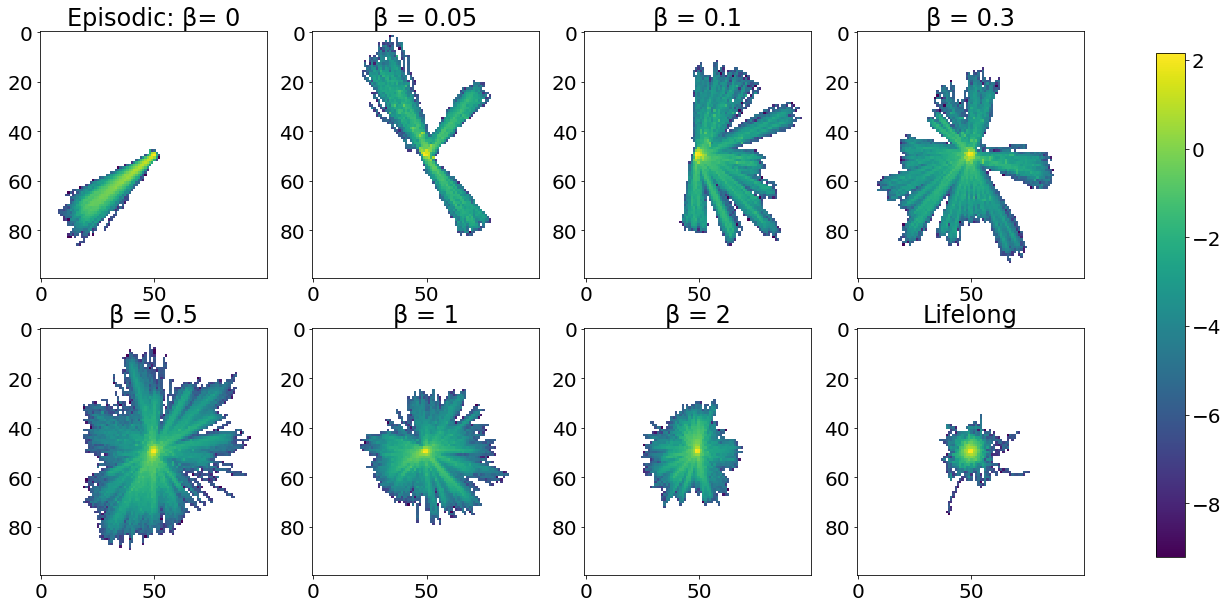}
\caption{State visitation on Ant during training under different values of $\beta$, colored by logarithmic visitation frequency (lighter = more visits).}
\label{fig:ablation}
\end{figure}
\textbf{Ablation on $\beta$.}
We vary $\beta$ on Ant with random starts near $(0,0)$ (Fig.~\ref{fig:ablation}). Increasing $\beta$ shrinks the exploration radius, confirming that a stronger lifelong reward constrains deep exploration in any one direction, but broadens angular coverage; a smaller $\beta$ reaches farther from the start. A balanced $\beta=0.5$ combines both effects.

\section{Conclusion}
\label{sec:conclusion}
We presented ELEMENT, an intrinsically motivated exploration framework that jointly maximizes episodic and lifelong state entropy: the episodic term searches deeper within episodes, the lifelong term searches wider across episodes, and both are computed non-parametrically from popular entropy estimators. ELEMENT achieves the highest episodic entropy and the largest state coverage among all compared methods, and learns diverse motion skills in a fully unsupervised manner that transfer to downstream tasks. Two limitations motivate future work: reliance on a fixed pre-trained encoder, which is hard to obtain in long-horizon environments, and the episode-length trade-off implied by Proposition~\ref{pro:optimal}.

\clearpage
\onecolumn
\section*{Supplementary Material}
\noindent This supplementary material accompanies the main paper and is organized as follows:
Sec.~\ref{ap:overview} (method overview and $k$NN-graph algorithms), Sec.~\ref{ap:related} (extended related work and baselines), Sec.~\ref{ap:proofs} (proofs of Propositions~1 and~2), Sec.~\ref{ap:details} (experimental details and hyper-parameters), and Sec.~\ref{ap:results} (episodic entropy curves and additional experimental results). Equation and proposition numbers in this document are local; references to ``Eq.~(x) of the main paper'' point to the main text.

\appendix
\section{Method overview and $k$NN-graph algorithms}\label{ap:overview}

\begin{figure}[htb]
\centering
\includegraphics[width=.55\linewidth]{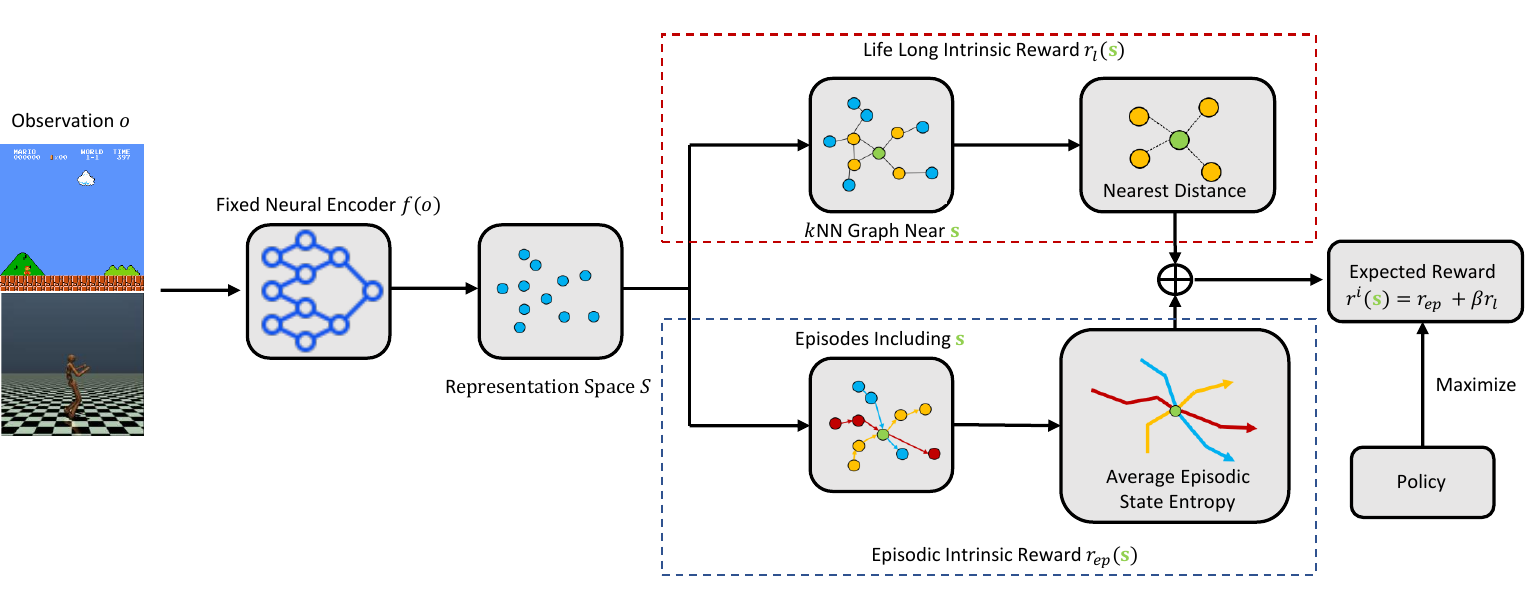}
\caption{Overview of ELEMENT. The lifelong intrinsic reward of a state $\mathbf{s}$ (green node) is the distance to its $k$-nearest neighbors, measured in a representation space obtained from a fixed neural encoder. The episodic intrinsic reward is the average state entropy of the episodes that include the state. The two rewards are combined and a deep RL agent is trained to maximize the expected reward.}
\label{fig:diagram}
\end{figure}

\begin{figure}[htb]
\centering
\includegraphics[width=.75\linewidth]{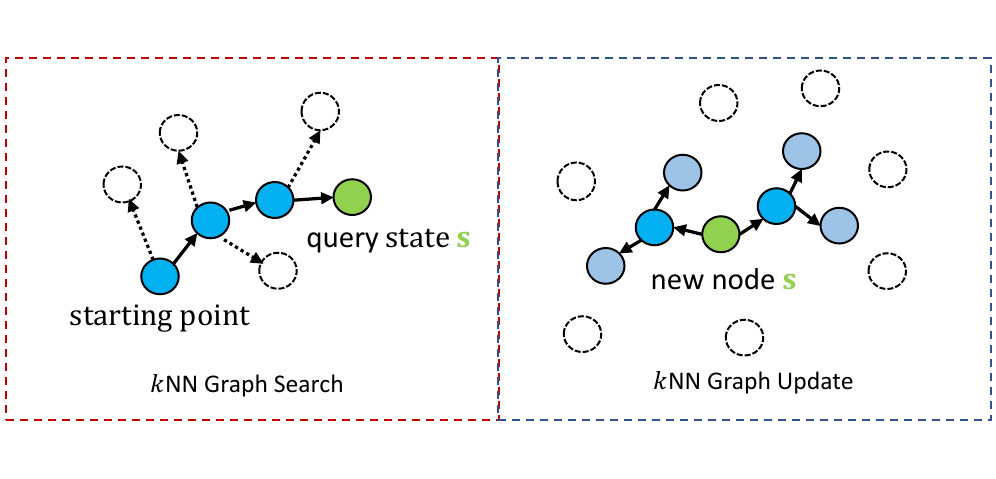}
\caption{$k$NN graph search and online update on simple nearest-neighbor graphs. Search moves to the neighbor closest to the query; the neighbors obtained in the search phase are then used as starting points to update existing nodes for which the new point is now a nearest neighbor.}
\label{fig:knn}
\end{figure}

Fig.~\ref{fig:diagram} gives an overview of the ELEMENT pipeline. Algorithms~\ref{alg:search} and~\ref{alg:update} detail the fast approximate $k$NN graph search~\cite{hajebi2011fast} used to compute the lifelong reward (Eq.~(\ref{eq:gnn-reward})) and the fast online graph update~\cite{debatty2016fast}; Fig.~\ref{fig:knn} illustrates both on simple nearest-neighbor graphs. In the search phase, starting from $R_2$ randomly chosen nodes, the algorithm repeatedly replaces the current node with the neighbor closest to the query, for at most $R_1$ steps. In the update phase, the neighbors found in the search phase are used as starting points to locate, up to depth $h$, existing nodes for which the new point is now a nearest neighbor.

\begin{figure}[htb]
\begin{minipage}[t]{0.48\linewidth}
\begin{algorithm}[H]
\caption{Fast Approximate $k$NN Graph Search}
\label{alg:search}
\begin{algorithmic}[1]
\REQUIRE $\mathcal{G}$: $k$NN graph; $\mathbf{s}$: query; $R_1$: greedy steps; $R_2$: restarts; $\text{ND}(x)$: $k$ neighbors of $x$; $d(\cdot,\cdot)$: distance.
\STATE $E\leftarrow\{\}$
\FOR{$r=1,\dots,R_2$ \textbf{(in parallel)}}
  \STATE Sample $x\sim\mathcal{G}$
  \FOR{$t=1,\dots,R_1$}
    \STATE $\mathcal{Y}\leftarrow\text{ND}(x)$;\quad $E\leftarrow E\cup\mathcal{Y}\cup\{x\}$
    \STATE $x\leftarrow\mathrm{argmin}_{y\in\mathcal{Y}\cup\{x\}}\,d(y,\mathbf{s})$
    \IF{$x$ unchanged}
      \STATE \textbf{break} \COMMENT{local minimum reached}
    \ENDIF
  \ENDFOR
\ENDFOR
\RETURN $k$ closest nodes to $\mathbf{s}$ in $E$
\end{algorithmic}
\end{algorithm}
\end{minipage}
\hfill
\begin{minipage}[t]{0.48\linewidth}
\begin{algorithm}[H]
\caption{Fast $k$NN Graph Online Update}
\label{alg:update}
\begin{algorithmic}[1]
\REQUIRE $\mathcal{G}$: $k$NN graph; $\mathbf{s}$: new node; $A_0$: $k$ neighbors of $\mathbf{s}$ (Alg.~\ref{alg:search}); $h$: update depth; $\text{ND}(x)$: $k$ neighbors of $x$; $d(\cdot,\cdot)$: distance.
\STATE Add $\mathbf{s}$ to $\mathcal{G}$ with edges to $A_0$;\quad $V\leftarrow A_0$
\FOR{$\ell=0,\dots,h-1$}
  \STATE $A_{\ell+1}\leftarrow\{\}$
  \FOR{each $x\in A_{\ell}$}
    \STATE $\mathcal{Y}\leftarrow\text{ND}(x)$;\quad $A_{\ell+1}\leftarrow A_{\ell+1}\cup(\mathcal{Y}\setminus V)$
    \IF{$d(x,\mathbf{s})<\max_{y\in\mathcal{Y}}d(x,y)$}
      \STATE Replace longest edge of $x$ with $(x,\mathbf{s})$
    \ENDIF
  \ENDFOR
  \STATE $V\leftarrow V\cup A_{\ell+1}$
\ENDFOR
\RETURN Updated $\mathcal{G}$
\end{algorithmic}
\end{algorithm}
\end{minipage}
\end{figure}

\section{Extended Related Work}\label{ap:related}

\textbf{State entropy maximization (non-parametric) for exploration.}
The Maximum Entropy (MaxEnt) principle is widely adopted in the exploration literature, augmenting the expected reward objective with an entropy term. Early works such as Soft Q-learning and SAC~\cite{haarnoja2017reinforcement,haarnoja2018soft} maximize the policy entropy $H(\pi(\mathbf{a}|\mathbf{s}))$ jointly with the extrinsic reward, yielding provably optimal and robust policies. However, these methods~\cite{haarnoja2018soft2,yang2021wcsac,eysenbach2021maximum} cannot drive exploration in the absence of extrinsic rewards. A separate line of work therefore maximizes a convex entropy function over the state visitation distribution, i.e., $H(\mathbf{s})$.

MaxEnt exploration methods fall into two categories. The first targets sampling efficiency, iteratively sampling states under the current policy to obtain entropy-related reward functions and solving the resulting MDP. Maximum Visitation Entropy Exploration (MVEE)~\cite{hazan2019provably} offers a provably efficient algorithm using mixed APPROXPLAN oracles that decompose state entropy into per-state gradients $\hat{r}(\mathbf{s})=\nabla H(\mathbf{s})=1/\hat{p}(\mathbf{s})$. (Hazan et al. named their approach MaxEnt; since this term is more commonly associated with policy entropy maximization~\cite{haarnoja2017reinforcement,eysenbach2021maximum}, we use MVEE.) Its R\'enyi variant~\cite{zhang2021exploration} uses the modified reward $\hat{r}(\mathbf{s})=\hat{p}^{(\alpha-1)}(\mathbf{s})$. Further improvements integrate representation learning~\cite{nedergaard2022k,yarats2021reinforcement} and reduce sample complexity~\cite{tiapkin2023fast}. However, sampling-based methods assume near-optimal MDP solutions at each iteration, which is impractical in complex environments.

The second category, to which ELEMENT belongs, develops intrinsic reward functions as substitutes for extrinsic rewards, enabling integration with any standard RL algorithm. These methods design dense per-state rewards $\hat{r}(\mathbf{s})$ whose sum over the collected memory is proportional to the state entropy. $k$NN-based methods~\cite{liu2021behavior,seo2021state,yuan2022renyi} define rewards proportional to the Kozachenko-Leonenko $k$NN estimator~\cite{singh2003nearest}: APT~\cite{liu2021behavior} uses $k$NN distances in a contrastive representation space, RE3~\cite{seo2021state} replaces the contrastive encoder with a fixed random encoder, and RISE~\cite{yuan2022renyi} extends the framework to R\'enyi entropy.

\textbf{Parametric methods for exploration.} (This extends the discussion in Sec.~\ref{sec:intro} of the main text.)
Parametric methods~\cite{pathak2017curiosity,ecoffet2019go,burda2018exploration,badia2020never,dewan2024curiosity} use an internal predictive model to measure novelty via prediction error, assigning larger rewards to less frequently visited states. These curiosity-driven approaches trace back to Pfaffelhuber's ``observer's information''~\cite{pfaffelhuber1972learning} and Lenat's ``interestingness''~\cite{lenat1976artificial}, and have resurged with deep neural networks. Modern instantiations differ in their internal model: ICM~\cite{pathak2017curiosity} and RND~\cite{burda2018exploration} predict the next observation with a CNN, while GIRIL~\cite{yu2020intrinsic} uses a variational autoencoder to model transitions. To mitigate vanishing novelty, NGU~\cite{badia2020never} and Go-Explore~\cite{ecoffet2019go} introduce episodic mechanisms. However, these approaches primarily augment extrinsic rewards, leaving pure exploration relatively underexplored; ELEMENT addresses this gap through state entropy maximization.

\textbf{Baselines.} RND~\cite{burda2018exploration} is a curiosity-driven parametric method that measures novelty via the prediction error of a fixed, randomly initialized network. NGU~\cite{badia2020never} combines lifelong novelty (prediction error) with an episodic memory that encourages within-episode state diversity. RE3~\cite{seo2021state} builds on APT~\cite{liu2021behavior}, replacing its contrastive encoder with a fixed random encoder. RISE~\cite{yuan2022renyi} extends RE3 to R\'enyi entropy using autoencoders and $k$NN estimation in the latent space; we use only its intrinsic reward, disabling the extrinsic component.

\section{Proofs}\label{ap:proofs}

\subsection{Proof of Proposition~\ref{pro:optimal}}\label{proof:optimal}

\begin{propositionrestatement}[Restatement of Proposition~\ref{pro:optimal}]
Assume all episodes in $D$ have the same length $T$ and each state occurs at most once per episode. Define the Monte-Carlo upper bound~\cite{ren2021learning}
\begin{equation}
\tilde{\mathcal{L}}(r_{ep})=\mathcal{L}(r_{ep})+\mathbb{E}_{\tau\in D}\big[T^2\,V_{\mathbf{s}\in\tau}(r_{ep}(\mathbf{s}))\big]\geq\mathcal{L}(r_{ep}),
\end{equation}
where $V_{\mathbf{s}\in\tau}(r_{ep}(\mathbf{s}))\geq0$ is the within-episode variance of the rewards and $\mathcal{L}$ is Eq.~(\ref{eq:loss}). The unique minimizer of $\tilde{\mathcal{L}}$ is
\begin{equation}\label{eq:optimal-app}
r^{*}_{ep}(\mathbf{s})=\tfrac{1}{T}\,\mathbb{E}_{\tau\in D}\big[H_{\mathbf{s}_i\in\tau}(\mathbf{s}_i)\,\big|\,\mathbf{s}\in\tau\big].
\end{equation}
\end{propositionrestatement}

The proof is inspired by trajectory-wise reward redistribution~\cite{gangwani2020learning,ren2021learning}.

\begin{proof}
For an episode $\tau$ and a time index $t\sim\mathcal{U}\{0,\dots,T-1\}$ sampled uniformly, define
\begin{equation}
\begin{aligned}
A&=H_{\mathbf{s}\in\tau}(\mathbf{s})-\textstyle\sum_{t'=0}^{T-1}r_{ep}(\mathbf{s}_{t'+1}),\\
B_t&=\textstyle\sum_{t'=0}^{T-1}r_{ep}(\mathbf{s}_{t'+1})-T\,r_{ep}(\mathbf{s}_{t+1}).
\end{aligned}
\end{equation}
$A$ does not depend on $t$, $\mathbb{E}_{t\sim\mathcal{U}}[B_t]=0$, and
\begin{equation}
\mathbb{E}_{t\sim\mathcal{U}}[B_t^2]=T^2\,\mathbb{E}_{t\sim\mathcal{U}}\big[(\bar r_\tau-r_{ep}(\mathbf{s}_{t+1}))^2\big]=T^2\,V_{\mathbf{s}\in\tau}(r_{ep}(\mathbf{s})),
\end{equation}
where $\bar r_\tau=\frac1T\sum_{t'}r_{ep}(\mathbf{s}_{t'+1})$ is the within-episode mean reward. Therefore,
\begin{equation}
\begin{aligned}
\tilde{\mathcal{L}}(r_{ep})
&=\mathcal{L}(r_{ep})+\mathbb{E}_{\tau\in D}\big[T^2\,V_{\mathbf{s}\in\tau}(r_{ep}(\mathbf{s}))\big]\\
&=\mathbb{E}_{\tau\in D}\big[\mathbb{E}_{t}[A^2]\big]+2\,\mathbb{E}_{\tau\in D}\big[A\,\mathbb{E}_{t}[B_t]\big]+\mathbb{E}_{\tau\in D}\big[\mathbb{E}_{t}[B_t^2]\big]\\
&=\mathbb{E}_{\tau\in D}\big[\mathbb{E}_{t}[(A+B_t)^2]\big]\\
&=\mathbb{E}_{\tau\in D}\Big[\mathbb{E}_{t}\big[(H_{\mathbf{s}\in\tau}(\mathbf{s})-T\,r_{ep}(\mathbf{s}_{t+1}))^2\big]\Big].
\end{aligned}
\end{equation}
Since $V_{\mathbf{s}\in\tau}(r_{ep})\geq0$, this also shows $\tilde{\mathcal{L}}(r_{ep})\geq\mathcal{L}(r_{ep})$.

To minimize $\tilde{\mathcal{L}}$, fix a state value $\mathbf{s}$ and collect the terms of the double expectation in which $\mathbf{s}_{t+1}=\mathbf{s}$. Because each state occurs at most once per episode, every episode $\tau$ containing $\mathbf{s}$ contributes exactly one such term, with weight $1/T$ from the uniform sampling of $t$. The objective is separable across state values, so for each $\mathbf{s}$ we minimize
\begin{equation}
g(r)=\mathbb{E}_{\tau\in D}\big[(H_{\mathbf{s}_i\in\tau}(\mathbf{s}_i)-Tr)^2\,\mathbb{I}(\mathbf{s}\in\tau)\big].
\end{equation}
Setting $g'(r)=-2T\,\mathbb{E}_\tau[(H_\tau-Tr)\mathbb{I}(\mathbf{s}\in\tau)]=0$ yields
\begin{equation}
r^{*}_{ep}(\mathbf{s})=\frac{\mathbb{E}_{\tau\in D}[H_{\mathbf{s}_i\in\tau}(\mathbf{s}_i)\,\mathbb{I}(\mathbf{s}\in\tau)]}{T\,\mathbb{E}_{\tau\in D}[\mathbb{I}(\mathbf{s}\in\tau)]}
=\tfrac1T\,\mathbb{E}_{\tau\in D}\big[H_{\mathbf{s}_i\in\tau}(\mathbf{s}_i)\,\big|\,\mathbf{s}\in\tau\big],
\end{equation}
which is Eq.~(\ref{eq:optimal-app}). Uniqueness follows from strict convexity of $g$ in $r$.
\end{proof}

\textbf{Varying episode lengths.} When episode lengths $T_\tau$ differ, the bound depends on how episodes are weighted. We define $\tilde{\mathcal{L}}$ as an expectation over transitions sampled uniformly from the buffer, matching line 14 of Algorithm~\ref{alg:element}, in which an episode is drawn with probability proportional to $T_\tau$:
\begin{equation}
\tilde{\mathcal{L}}(r_{ep})=\frac{1}{\sum_{\tau\in D}T_\tau}\sum_{\tau\in D}\sum_{t=0}^{T_\tau-1}\big(H_{\mathbf{s}\in\tau}(\mathbf{s})-T_\tau\,r_{ep}(\mathbf{s}_{t+1})\big)^2.
\end{equation}
Repeating the argument above (the per-state objective becomes $g(r)=\sum_{\tau\in D}T_\tau\,(H_{\mathbf{s}_i\in\tau}(\mathbf{s}_i)-T_\tau r)^2\,\mathbb{I}(\mathbf{s}\in\tau)$, and setting $g'(r)=0$), the minimizer becomes
\begin{equation}
r^{*}_{ep}(\mathbf{s})=\frac{\sum_{\tau\in D}T_\tau\,H_{\mathbf{s}_i\in\tau}(\mathbf{s}_i)\,\mathbb{I}(\mathbf{s}\in\tau)}{\sum_{\tau\in D}T_\tau^2\,\mathbb{I}(\mathbf{s}\in\tau)},
\end{equation}
This length-weighted form is the one referred to in Sec.~\ref{sec:episodic} of the main text.

\subsection{Proof of Proposition~\ref{pro:intrinsic}}\label{proof:intrinsic}

\begin{propositionrestatement}[Restatement of Proposition~\ref{pro:intrinsic}]
Given $N$ states $\{\mathbf{s}_i\}_{i=1}^{N}$ and a Gaussian kernel $\kappa_\sigma$, let $\knn(i)$ denote the index set of the $k$-nearest neighbors of $\mathbf{s}_i$ (excluding $\mathbf{s}_i$ itself), $\mathbf{s}_i^{(k)}$ its $k$-th nearest neighbor, and $\rho_i=\|\mathbf{s}_i-\mathbf{s}_i^{(k)}\|_2$. Fix a tolerance $\varepsilon\in(0,1)$. Then:
\begin{align}
\text{(i)}\;\; & H_{\knn}(\mathbf{s})=\tfrac{d}{N}\textstyle\sum_{i=1}^{N}\log\rho_i+c_{N,k,d},\quad c_{N,k,d}=\log N+\log\tfrac{\pi^{d/2}}{k\,\Gamma(d/2+1)}+C_k,\notag\\
\text{(ii)}\;\; & \Big|H_{\text{KDE}}(\mathbf{s})-\Big(\log N-\tfrac1N\textstyle\sum_{i=1}^{N}\log\big(1+\sum_{j\in\knn(i)}\kappa(\mathbf{s}_i,\mathbf{s}_j)\big)\Big)\Big|\leq\varepsilon\notag\\
& \quad\text{if }\forall i,\;\rho_i\geq\big[2\sigma\log\tfrac{N-k-1}{\varepsilon}\big]^{1/2},\notag\\
\text{(iii)}\;\; & H_2(\mathbf{s})=2\log_2N-\log_2\Big(N+\textstyle\sum_{i=1}^{N}\sum_{j\in\knn(i)}\kappa^2(\mathbf{s}_i,\mathbf{s}_j)\Big)+\delta,\;\;|\delta|\leq\varepsilon/\ln2,\notag\\
& \quad\text{if }\forall i,\;\rho_i\geq\big[\sigma\log\tfrac{N-k-1}{\varepsilon}\big]^{1/2}.\notag
\end{align}
Relation (i) is exact and requires no condition. In (ii) and (iii), the truncation exploits the exponential decay of the Gaussian kernel: for any $\mathbf{s}_m$ far from $\mathbf{s}_i$, $\kappa_\sigma(\mathbf{s}_i,\mathbf{s}_m)\to0$. Since $\kappa^2$ has effective bandwidth $\sigma/2$, the condition in (iii) is milder than that in (ii) by a factor of $\sqrt2$.
\end{propositionrestatement}

\begin{proof}
\textbf{(i) $k$NN estimator.} By definition (Eq.~(\ref{eq:knn})),
\begin{equation}
H_{\knn}(\mathbf{s})=\frac1N\sum_{i=1}^{N}\log\frac{N\rho_i^{d}\pi^{d/2}}{k\,\Gamma(d/2+1)}+C_k
=\frac dN\sum_{i=1}^{N}\log\rho_i+\underbrace{\log N+\log\frac{\pi^{d/2}}{k\,\Gamma(d/2+1)}+C_k}_{=\,c_{N,k,d}},
\end{equation}
which is exact and requires no condition.

\textbf{(ii) Kernel density estimator.} Since $\kappa(\mathbf{s}_i,\mathbf{s}_i)=1$, Eq.~(\ref{eq:kde}) becomes
\begin{equation}
H_{\text{KDE}}(\mathbf{s})=-\frac1N\sum_{i=1}^{N}\log\Big[\frac1N\Big(1+\sum_{j\neq i}\kappa(\mathbf{s}_i,\mathbf{s}_j)\Big)\Big]
=\log N-\frac1N\sum_{i=1}^{N}\log(1+S_i+R_i),
\end{equation}
where $S_i=\sum_{j\in\knn(i)}\kappa(\mathbf{s}_i,\mathbf{s}_j)$ and $R_i=\sum_{j\notin\knn(i),\,j\neq i}\kappa(\mathbf{s}_i,\mathbf{s}_j)$ is the truncated tail. Sorting the remaining $N-k-1$ points by distance, every term of $R_i$ is at most $\exp(-\rho_i^2/2\sigma)$, hence
\begin{equation}
0\leq R_i\leq(N-k-1)\,e^{-\rho_i^2/2\sigma}\leq\varepsilon
\quad\text{whenever}\quad
\rho_i\geq\big[2\sigma\log\tfrac{N-k-1}{\varepsilon}\big]^{1/2}.
\end{equation}
Since the argument of the logarithm is at least 1,
\begin{equation}
0\leq\log(1+S_i+R_i)-\log(1+S_i)=\log\Big(1+\frac{R_i}{1+S_i}\Big)\leq\frac{R_i}{1+S_i}\leq R_i\leq\varepsilon.
\end{equation}
Averaging over $i$ yields the claimed bound.

\textbf{(iii) Second-order matrix-based R\'enyi estimator.} Since $\kappa(\mathbf{s}_i,\mathbf{s}_i)=1$, $\tr(K)=N$ exactly, so $A=K/N$ and $\tr(A^2)=\frac{1}{N^2}\sum_{i}\sum_{j}\kappa^2(\mathbf{s}_i,\mathbf{s}_j)$. Hence
\begin{equation}
H_2(\mathbf{s})=\frac{1}{1-2}\log_2\tr(A^2)=2\log_2N-\log_2\Big(\underbrace{N+\sum_{i=1}^{N}\sum_{j\neq i}\kappa^2(\mathbf{s}_i,\mathbf{s}_j)}_{=:X}\Big),
\end{equation}
where the diagonal contributes $\sum_i\kappa^2(\mathbf{s}_i,\mathbf{s}_i)=N$. Write $X=N+\sum_i(S'_i+R'_i)$ with $S'_i=\sum_{j\in\knn(i)}\kappa^2(\mathbf{s}_i,\mathbf{s}_j)$ and truncated tail $R'_i\geq0$. Because $\kappa^2(\mathbf{s}_i,\mathbf{s}_j)=\exp(-\|\mathbf{s}_i-\mathbf{s}_j\|_2^2/\sigma)$ is a Gaussian kernel of bandwidth $\sigma/2$, the same argument as in (ii) gives
\begin{equation}
0\leq R'_i\leq(N-k-1)\,e^{-\rho_i^2/\sigma}\leq\varepsilon
\quad\text{whenever}\quad
\rho_i\geq\big[\sigma\log\tfrac{N-k-1}{\varepsilon}\big]^{1/2},
\end{equation}
a condition milder than that of (ii) by a factor of $\sqrt2$. Let $\tilde X=N+\sum_iS'_i\geq N$ denote the truncated sum, so that $0\leq X-\tilde X\leq N\varepsilon$. By the mean value theorem,
\begin{equation}
0\leq\log_2X-\log_2\tilde X\leq\frac{X-\tilde X}{\tilde X\ln2}\leq\frac{N\varepsilon}{N\ln2}=\frac{\varepsilon}{\ln2},
\end{equation}
which establishes the claim with $|\delta|\leq\varepsilon/\ln2$. Finally, $H_2$ is a strictly decreasing function of $\sum_i\sum_{j\in\knn(i)}\kappa^2(\mathbf{s}_i,\mathbf{s}_j)$, which yields the monotonicity property discussed after Proposition~\ref{pro:intrinsic}, but not an additive decomposition.
\end{proof}

\section{Details of Experimental Setting}\label{ap:details}
All experiments are conducted on a single V100 GPU per training process, with a maximum memory usage of 10GB.

\subsection{State space for entropy maximization}\label{ap:state_space}
We adopt fixed neural encoders $f(o)$ to map raw observations to a state space suitable for entropy estimation. For the MuJoCo environments, we use fully connected random networks with two hidden layers of 64 neurons each, encoding observations into 5-dimensional vectors, which are then concatenated with the spatial coordinates and torso orientations. State entropy is maximized over these concatenated representations. For Mario, we pre-train an encoder using Spatiotemporal Deep Infomax~\cite{anand2019unsupervised}, a contrastive method that maximizes mutual information across spatial and temporal axes, mapping game frames to a 32-dimensional representation space. Hyper-parameters are provided in Table~\ref{tab:encoder}.

\begin{table}[htb]
\centering\small
\begin{tabular}{lc}
\toprule
Hyper-parameter & Value \\ \midrule
number of steps & 5e+6 \\
num-processes & 8 \\
lr & 3e-4 \\
batch-size & 64 \\
use\_multiple\_predictors & False \\
patience & 15 \\
feature-size & 32 \\
entropy-threshold & 0.6 \\
layer\_num & 2 \\
activation\_function & relu \\
last\_activation & None \\
collect-mode & random\_agent \\
\bottomrule
\end{tabular}
\caption{Hyper-parameters of Spatiotemporal Deep Infomax.}
\label{tab:encoder}
\end{table}

\subsection{Hyper-parameter setting of ELEMENT}\label{ap:hyper_element}
Hyper-parameter settings for all five environments are provided in Table~\ref{tab:element}. Note that the episodic module on Walker2D, Ant, and Humanoid uses the matrix-based estimator with $\alpha=3$, which lies outside the $\alpha=2$ scope of Proposition~\ref{pro:intrinsic}; this is immaterial because the episodic module computes whole-episode entropy directly via Eq.~(\ref{eq:renyi}) and never relies on the $k$NN decomposition, which is used only by the lifelong module.

\begin{table}[htb]
\centering\small
\begin{tabular}{lccccc}
\toprule
Hyper-parameter & Hopper & Walker2D & Ant & Humanoid & Mario \\ \midrule
$\beta$ & 0.5 & 0.5 & 0.5 & 0.5 & 0.5 \\
estimator & KDE & Matrix-based R\'enyi & Matrix-based R\'enyi & Matrix-based R\'enyi & $k$NN \\
$\sigma$ & 1 & 1 & 1 & 1 & -- \\
$\alpha$ & -- & 3 & 3 & 3 & -- \\
$k$\_estimator & -- & -- & -- & -- & 5 \\
$T_\tau$ & 1000 & 1000 & 1000 & 1000 & 300 \\
graph\_search\_step ($R_1$) & 20 & 20 & 20 & 20 & 10 \\
num\_graph\_search\_initial ($R_2$) & 20 & 20 & 10 & 10 & 5 \\
$h$ (graph update depth) & 2 & 2 & 2 & 2 & 2 \\
$k$\_lifelong & 3 & 3 & 3 & 3 & 3 \\
$U$ & 5e+5 & 5e+5 & 5e+5 & 5e+5 & 1e+5 \\
$T_u$ & 5e+4 & 5e+4 & 5e+4 & 5e+4 & 2e+4 \\
\bottomrule
\end{tabular}
\caption{Hyper-parameters of ELEMENT on the five environments.}
\label{tab:element}
\end{table}

\subsection{Hyper-parameter setting of RL agents}\label{ap:hyper_rl}
For the MuJoCo environments, we use SAC\footnote{\url{https://github.com/seolhokim/Mujoco-Pytorch}} as the backbone RL algorithm, with hyper-parameters summarized in Table~\ref{tab:sac}. For Mario, we use A3C as the backbone, with hyper-parameters in Table~\ref{tab:a3c}. The A3C network architecture is:
\begin{equation}
\begin{split}
&self.conv1 = nn.Conv2d(1, 32, 3, stride=2, padding=1)\\
&self.conv2 = nn.Conv2d(32, 32, 3, stride=2, padding=1)\\
&self.conv3 = nn.Conv2d(32, 32, 3, stride=2, padding=1)\\
&self.conv4 = nn.Conv2d(32, 32, 3, stride=2, padding=1)\\
&self.lstm = nn.LSTMCell(32 * 6 * 6, 512)\\
&self.critic\_linear = nn.Linear(512, 1)\\
&self.actor\_linear = nn.Linear(512, 16)
\end{split}
\end{equation}

\begin{table}[htb]
\centering\small
\begin{tabular}{lc}
\toprule
Hyper-parameter & Value \\ \midrule
initial temperature & 0.2 \\
gamma & 0.99 \\
actor\_lr & 3e-4 \\
critic\_lr & 3e-4 \\
q\_lr & 3e-4 \\
soft\_update\_rate & 0.005 \\
hidden\_dim & 256 \\
memory size & 1e+6 \\
layer\_num & 3 \\
batch\_size & 128 \\
activation\_function & torch.relu \\
last\_activation & None \\
\bottomrule
\end{tabular}
\caption{Hyper-parameters of SAC.}
\label{tab:sac}
\end{table}

\begin{table}[htb]
\centering\small
\begin{tabular}{lc}
\toprule
Hyper-parameter & Value \\ \midrule
action\_type & complex \\
gamma & 0.9 \\
actor\_lr & 1e-4 \\
critic\_lr & 1e-4 \\
parameter for GAE & 1.0 \\
entropy coefficient & 0.01 \\
num\_local\_steps & 300 \\
num\_processes & 6 \\
activation\_function & torch.relu \\
last\_activation & None \\
\bottomrule
\end{tabular}
\caption{Hyper-parameters of A3C.}
\label{tab:a3c}
\end{table}

\section{Additional Experimental Results}\label{ap:results}

\subsection{Episodic state entropy over training}\label{ap:ep_curves}
\begin{figure}[!htb]
\centering
\includegraphics[width=.9\linewidth]{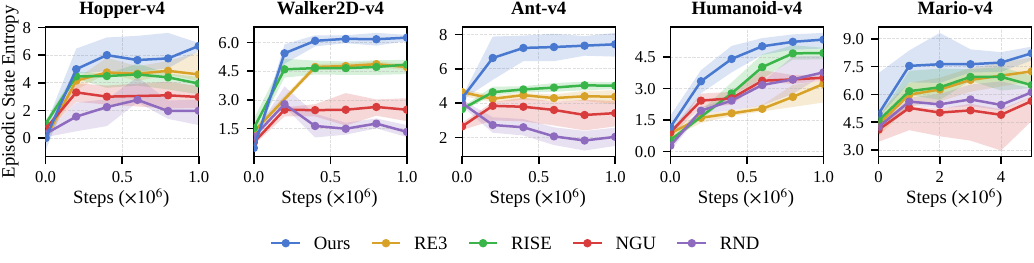}
\caption{Episodic state entropy throughout training, with $\beta=0$ to isolate the episodic module. The $y$-axis is the within-episode Shannon entropy estimated with the matrix-based functional~\cite{giraldo2014measures} at $\alpha=1.001$; shaded regions denote one standard deviation, mean over 10 runs.}
\label{fig:ep_curves}
\end{figure}
Table~\ref{tab:final_entropy} of the main text reports the entropy reached at the final checkpoint. Fig.~\ref{fig:ep_curves} gives the full learning curves behind those values. ELEMENT not only ends higher in every environment but separates from the baselines within the first few hundred thousand steps and holds the gap thereafter, so the advantage reflects faster exploration rather than a better final point alone.

\subsection{How does ELEMENT drive the agent?}\label{ap:policy}
\begin{figure}[htb]
\centering
\includegraphics[width=.8\linewidth]{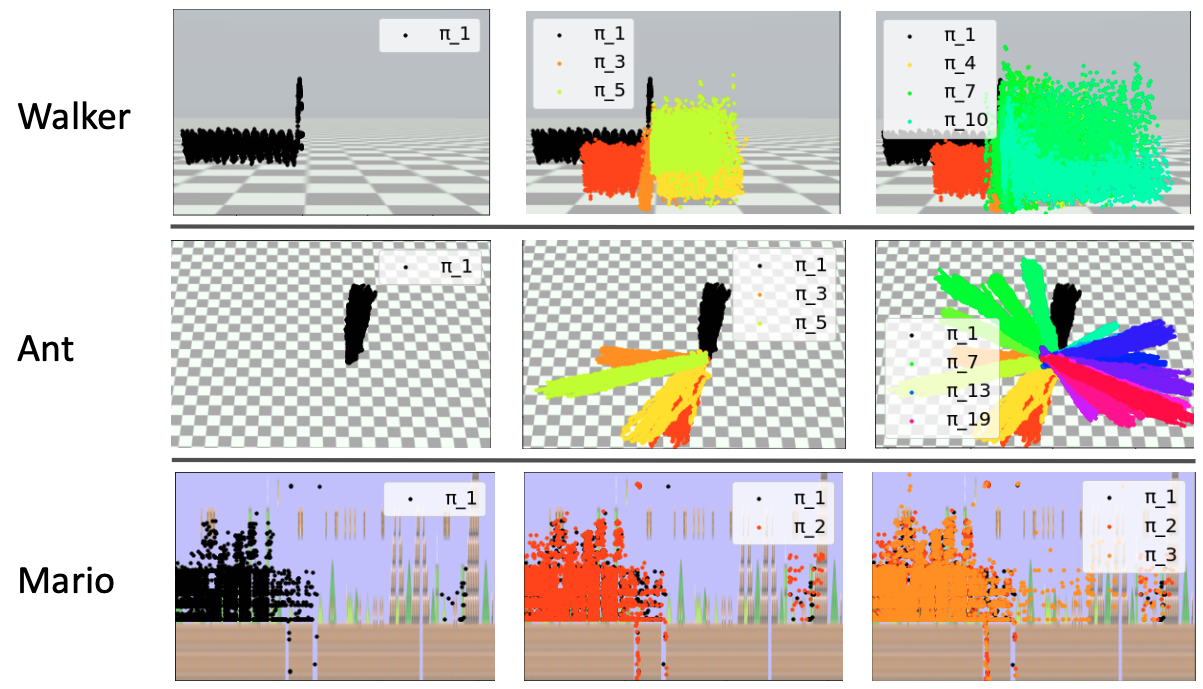}
\caption{Trajectories collected at different training checkpoints (different colors) on Walker, Ant, and Mario. Initially, the episodic reward drives the agent to produce trajectories whose states are diverse within each episode but similar across episodes. As the $k$NN graph is updated, the lifelong reward gradually steers the agent toward unvisited regions, while within-episode diversity is maintained.}
\label{fig:policy}
\end{figure}
Fig.~\ref{fig:policy} visualizes the exploration process on Walker, Ant, and Mario at checkpoints taken every $50$k steps, executing the checkpoint policy to collect $30$k samples per checkpoint. Early on, the episodic reward yields trajectories that are diverse within each episode but similar across episodes; as the $k$NN graph is updated, the lifelong reward steers the agent toward unvisited regions while within-episode diversity is maintained. Videos are available at \url{https://sites.google.com/view/element-rl}.

\subsection{How does ELEMENT distribute rewards?}\label{ap:maze}
We use a $20\times20$ discrete maze\footnote{\url{https://github.com/giorgionicoletti/deep_Q_learning_maze}} to visualize the intrinsic reward distributions during training. Agents start from the top-left corner and navigate for 700 steps per episode. We train separate agents using only $r_{ep}$ or only $r_l$ to isolate and compare their reward distributions. Since the maze is small, with a finite state space, we use brute-force $k$NN instead of a graph. We employ tabular Q-learning as the backbone, updated over $300\times700$ steps.

\begin{figure}[htb]
\centering
\begin{minipage}[b]{0.42\linewidth}
\centering
\centerline{\includegraphics[width=\linewidth]{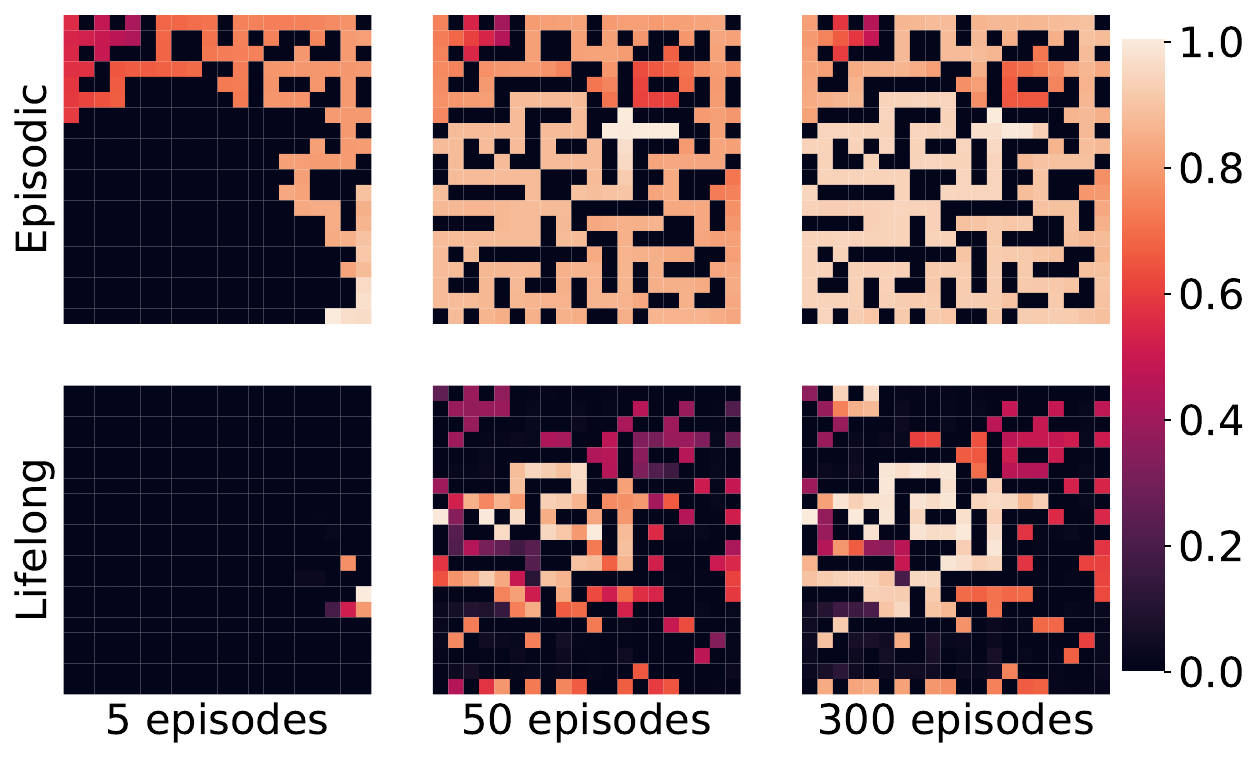}}
\centerline{Rewards}\medskip
\end{minipage}
\hspace{0.03\linewidth}
\begin{minipage}[b]{0.30\linewidth}
\centering
\centerline{\includegraphics[width=\linewidth]{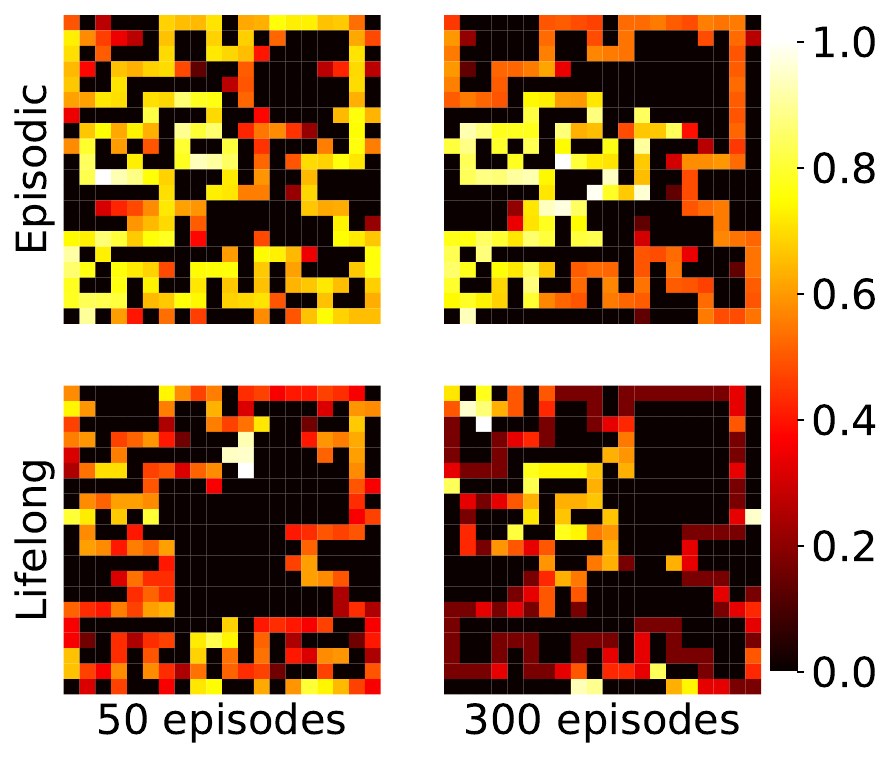}}
\centerline{Visited States}\medskip
\end{minipage}
\caption{Left: reward distributions assigned by the episodic and lifelong rewards at different training stages. Right: corresponding agent trajectories.}
\label{fig:maze}
\end{figure}

Fig.~\ref{fig:maze} shows reward snapshots at $5\times700$, $50\times700$, and $300\times700$ steps. Initially, the lifelong reward motivates the agent to explore new areas. By episode 50, it assigns high rewards to deeper states. However, as the agent reaches these deeper states more frequently, the rewards near the starting point gradually increase while the rewards along the paths to the deeper states decrease. By episode 300, the agent tends to become trapped near the starting point. In contrast, the episodic reward distributes rewards smoothly across the maze, reaching even the deepest states. This highlights a key advantage of the episodic reward: it recognizes that frequently visited states near the start may serve as essential gateways to less explored regions, and therefore maintains high rewards along entire trajectories rather than concentrating them at novel endpoints.

\subsection{Can ELEMENT augment extrinsic rewards?}\label{ap:augment}
\begin{figure}[htb]
\centering
\includegraphics[width=.9\linewidth]{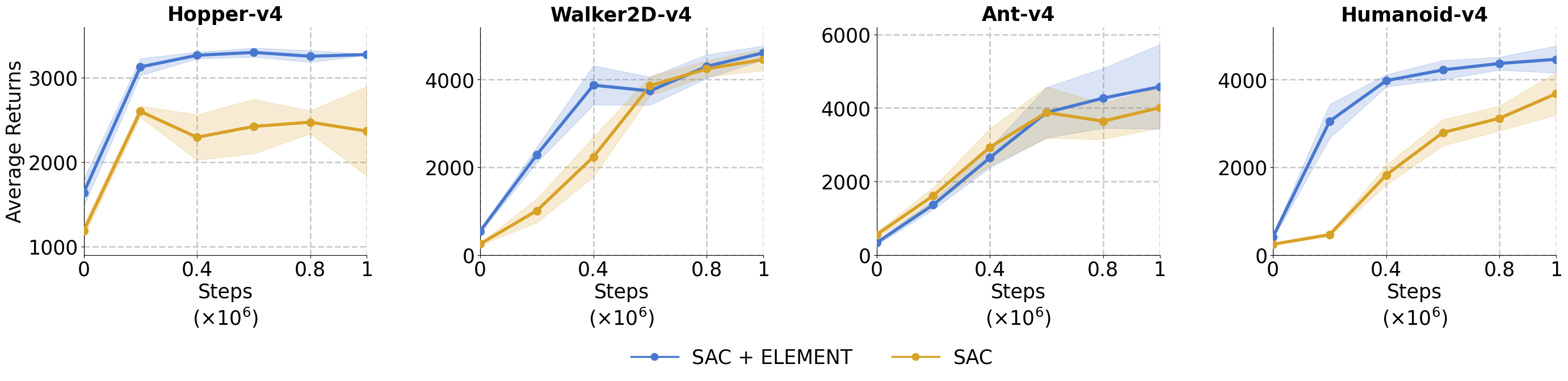}
\caption{Sample efficiency of SAC augmented with the ELEMENT episodic reward across MuJoCo environments.}
\label{fig:augment}
\end{figure}

We examine whether ELEMENT can improve sample efficiency when combined with extrinsic task rewards. The total reward is $r^{\text{total}}=r^{\text{Extrinsic}}+\lambda\cdot r^i$, where $\lambda$ balances the two components. We set $\lambda=0.1$ and $\beta=0$ to focus on the episodic reward $r_{ep}$, as the benefit of augmenting with the lifelong reward $r_l$ has already been validated by prior work~\cite{seo2021state,yuan2022renyi}. Results are averaged over 5 seeds with variance reported, and the learning curves in Fig.~\ref{fig:augment} are smoothed for clarity. ELEMENT consistently improves the sample efficiency of SAC across all environments.

\clearpage
\bibliographystyle{IEEEbib}
\bibliography{strings}

\end{document}